\documentclass[sigconf]{acmart}

\usepackage{cleveref}

\AtBeginDocument{%
  \providecommand\BibTeX{{%
    \normalfont B\kern-0.5em{\scshape i\kern-0.25em b}\kern-0.8em\TeX}}}

\setcopyright{acmcopyright}
\copyrightyear{2018}
\acmYear{2018}
\acmDOI{10.1145/1122445.1122456}

\acmConference[Woodstock '18]{Woodstock '18: ACM Symposium on Neural
  Gaze Detection}{June 03--05, 2018}{Woodstock, NY}
\acmBooktitle{Woodstock '18: ACM Symposium on Neural Gaze Detection,
  June 03--05, 2018, Woodstock, NY}
\acmPrice{15.00}
\acmISBN{978-1-4503-XXXX-X/18/06}

\begin{document}

\title{Generative Forensics: \\ Procedural Generation and Information Games}


\author{Michael Cook}
\email{mike@possibilityspace.org}
\affiliation{%
  \institution{School of Electrical Engineering and Computer Science\\
  Queen Mary University of London}
}

\renewcommand{\shortauthors}{Author}

\begin{abstract}

Procedural generation is used across game design to achieve a wide variety of ends, and has led to the creation of several game subgenres by injecting variance, surprise or unpredictability into otherwise static designs. Information games are a type of mystery game in which the player is tasked with gathering knowledge and developing an understanding of an event or system. Their reliance on player knowledge leaves them vulnerable to spoilers and hard to replay. In this paper we introduce the notion of \textit{generative forensics} games, a subgenre of information games that challenge the player to understand the output of a generative system. We introduce information games, show how generative forensics develops the idea, report on two prototype games we created, and evaluate our work on generative forensics so far from a player and a designer perspective.
%
\end{abstract}




\maketitle

\section{Introduction}
Procedural content generation (PCG), a technique where some game content is created by an algorithmic process, has been present in game development for over four decades \cite{elite, rogue}. It is used for a variety of purposes, including assisting with repetitive or at-scale content creation processes \cite{speedtree}, or introducing unpredictability into a game design \cite{spelunkybook}. While this has led to many misconceptions about the function of PCG, it has also enabled the creation of many genre variants in which traditionally static design elements are replaced with dynamic, unpredictable or variable content. 

Common justifications for the use of PCG in game design is that it increases replayability or reduces development cost or complexity \cite{searchbasedpcg}. Although these claims regularly appear in academic papers about PCG, to our knowledge there are no studies to support either claim, and in our experience adding PCG can actually \textit{increase} the complexity of the development process. A better justification for PCG is that it enables new kinds of game design, by changing the player's relationship with the game's systems, or allowing the developers to work at a scale that would be otherwise impractical. The best-known example of this is Spelunky, a blend of platformer and roguelike \cite{spelunky}. In his book of the same name \cite{spelunkybook}, designer Derek Yu explained that the use of procedural generation in Spelunky was intended to reduce the reliance on rote learning in platformers, by making levels unpredictable. Yu's use of procedural generation to disrupt rote learning became a trend throughout the 2010s. Although the role of PCG in these games is reduced in marketing terms to `replayability', PCG is serving a specific design purpose, making the player improvise each time they play.

An \textit{information game} is a game in which the player is tasked with understanding a complex artefact -- a past sequence of events, a language, a physical system -- by gaining knowledge about it, drawing inferences from this knowledge, and using this to seek out new knowledge \cite{francis}. A common theme in such games is a sense of mystery -- the quest to gain understanding is fundamental to the game's dramatic arc. This makes information games more fragile to player knowledge -- players may feel less inclined to play an information game twice, the game is more vulnerable to spoilers than most narrative games and players can accidentally `shortcut' the game and stumble on a solution unintentionally.

In this paper we propose a new variant subgenre of information games we call \textit{generative forensics games}, which use simulation-based PCG to create the mystery for the player to solve. This helps ease some of the problems mentioned above, such as a vulnerability to spoilers, but more importantly it opens up interesting new design possibilities related to generative design, storytelling, and player engagement with PCG systems. Our prototypes demonstrate our early exploratory work in the area, but have already yielded interesting responses from players and a lot of reflections as developers.

The remainder of this paper is organised as follows: in \cref{sec:background} we introduce information games and describe four canonical examples, and then in \cref{sec:infogames} we identify key features of such games. In \cref{sec:motivation} we connect information games to procedural content generation, and then in \cref{sec:nbr} and \cref{sec:cu} we describe and evaluate our two prototype generative forensics games: \textit{Nothing Beside Remains} and \textit{Condition Unknown}. In \cref{sec:fw} we discuss general future directions for the work, and then summarise in \cref{sec:conclusion}.
%
%
%

\section{Background - Information Games}\label{sec:background}
\textit{Information game} is a term coined by developer and critic Tom Francis in 2019. In \cite{francis} he defines an information game as follows:

\begin{quote}\textit{An information game is a game where the goal is to acquire information, and the way you do it is to use information you've \textit{already} gained and reason about it, deduce things from it, come up with theories and use those theories to go looking for more information.}\end{quote}

Francis cites \textit{Return of the Obra Dinn}, \textit{Heaven's Vault}, \textit{Her Story} and \textit{Outer Wilds} as examples of information games, and notes that it is not a genre, but rather a particular category of games that share a particular approach to player knowledge. In this section we will examine these four games and identify the aspects of their gameplay that embody this category of game. \textbf{Note that this paper includes light non-endgame spoilers for these games.}


\subsection{Return of the Obra Dinn}
Return of the Obra Dinn is a narrative mystery game in which you play an insurance adjuster sent aboard a lost ship (the titular \textit{Obra Dinn}) which has reappeared after many years lost at sea \cite{obradinn}. The player possesses a passenger manifest and a sketch of the ship's crew; they must identify every person in the sketch and determine their `fate' -- one of 24 causes of death, or \textit{alive}. Some fates require extra information to be supplied. For example, if someone is shot then the shooter must also be identified, or if someone is still alive then the country they fled to must be stated. To aid the player in determining this, they can examine dead bodies found on the ship and see a vision of that person's death. The player hears a few seconds of audio preceding the event, and then sees a frozen scene they can walk around, looking for clues as to what happened. Finding dead bodies within a past scene will unlock a nested scene. 

The game contains many systems and tools to make the player's job easier. During a death scene, the player can view the manifest and see which passengers from the drawing of the crew are visible in the current scene (although sometimes passengers may be concealed). The game also tells the player what approximate chronological order the scenes are in. There is a system to indicate problem difficulty to the player: each passenger is marked with a rating indicating how hard it will be to deduce their identity, and faces are blurred in the crew sketch if the game believes the player does not have enough information to identify them yet. Note that this blurring is based purely on which death scenes have been viewed and is hardcoded by the developer, as are the difficulty estimations. The game does not know what information the player has understood, just which scenes they have discovered.

Information can be gathered from scenes in different ways. Some evidence is more overt, such as an early death scene which features someone mentioning someone else by name before audibly attacking them. Other evidence is more subtle. For example, people sat playing cards together may know each other, which might suggest they work together. Some people are in formal uniforms in the crew sketch which helps infer their rank. The passenger manifest also lists the nationality of each crew member, which lets the player interpret accents and language clues in audio clips, as well as inferring which crew members are likely to talk to one another.

Any time the player's book contains three correctly identified passengers with their fate, the game immediately interrupts and confirms the three correct entries. This can help players test out theories as well as realise if they have put in too many incorrect fates. The number of correctly identified passenger fates also affects the type of ending received.

\subsection{Heaven's Vault}
Heaven's Vault is a linguistic puzzle game in which you play an archaeologist investigating an ancient civilisation \cite{heavensvault}. Heaven's Vault has adventure game elements where the player travels between locations, talks to characters, collects items and solves puzzles. However, its unique selling point is the ancient language the player encounters throughout the game. The player must translate the text they find by learning the language using context clues, knowledge about the world's history, and intuition from the shape of glyphs.  The game does not immediately confirm how accurate their translations are, which obfuscates the process and encourages hypothesis-forming. When a new piece of text is found it must be broken down into words (sometimes done automatically) and can then be translated. A word can be in three states: \textit{translated}, meaning the player has previously confirmed its translation; \textit{tentative}, meaning the player has encountered the word before and suggested a translation, but it has not been confirmed; and \textit{new}, meaning the player has never seen this word before.

\begin{figure}
\includegraphics[width=\columnwidth]{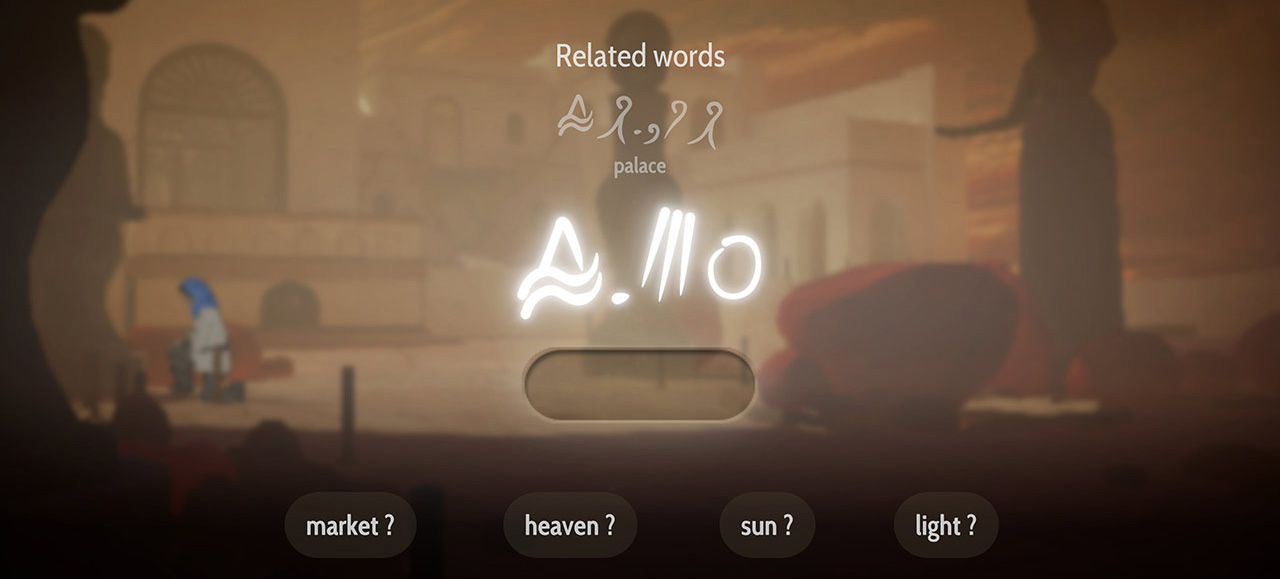}
\caption{Translating a word in \textit{Heaven's Vault}.}
\label{fig:hv}
\end{figure}

The game presents players with several possible translations for a given word. Although the language uses a similar sentence structure to English, the language differs in other key ways. The language is somewhat pictorial in nature, which the developers say was inspired by Ancient Egyptian and Chinese. Words are also compositional, similar to a language like German. This encourages players to guess at how smaller words relate to larger ones. For example, the word for \textit{home} is comprised of glyphs for \textit{place} and \textit{living}. Hints as to what a glyph or word means can come from the context in which it is found, the other words in the sentence, other known words that contain similar glyphs or simply looking at the glyphs themselves (e.g. wavy lines meaning water).%

Some words are translated at key plot points, but most confirmations come through repeated use. Once a word has been successfully used enough times the game will confirm the translation. However, the game's narrative structure is very flexible, and incorrect translations can be left unconfirmed for some time, leading the player down different narrative paths. This can lead to dramatic twists when the player discovers the true meaning of a word which recontextualises previous phrases. The game mixes in many nonessential texts alongside story-relevant pieces, which provide more context for words and more opportunities to produce confirmed translations. These nonessential texts are generated procedurally, and can dynamically increase in complexity as the game progresses, or on successive playthroughs of the game, adding new language elements and words. The developers estimate that players will still learn new language constructs on their fourth playthrough.

\subsection{Her Story}
Her Story is an FMV puzzle game in which the player searches through an archive of police interviews to solve a mystery \cite{herstory}. Players search the archive by typing in text queries, but results are sorted chronologically and only the earliest five results are shown. Additionally, the interviews are broken up into many small clips, usually only a few seconds long. This sets up an information-gathering challenge for the player: very common words appear many times; searching for words like ``murder'' will return very early video clips that the player has probably already seen. More specific words may appear in fewer clips, or later chronologically as the story develops, but because the search query is totally unconstrained, the player initially has no sense of what to search for.

The player can tag video clips to note information, and view a `database checker' which shows how many clips have been found and seen. The database checker is a series of boxes that are either checked or unchecked, and offer no information about the clip they refer to except that the boxes are arranged in chronological order. Initially the player is likely to search based on hypothetical ideas -- searching for \textit{weapon}, perhaps, or \textit{victim}. As important people or concepts are mentioned they enable the player to make more specific searches, which result in later video clips being surfaced.

Her Story has no prescribed ending. After the player has seen specific clips, a chat window appears on the interface and asks them two questions: `Are you done?' and, if the player responds with `yes', asks a second question which reveals a crucial piece of information. This second question is a significant plot revelation, and may encourage the user to go back and make new searches. The player can return to this chat window at any time -- answering `yes' to both questions ends the game.


Her Story's structure is built out of player knowledge. There is nothing to stop the player from typing in crucial search terms within minutes of starting the game and uncovering twists. However, the script is structured in such a way that the player slowly learns about new people, locations, dates and events, which inspire ideas for new queries. This open-endedness is vital in giving the game a sense of mystery-solving, but it also leaves the game open to being `short-circuited' by a hunch or good guess (searching for a reasonably common trope in crime mystery opens up a major line of inquiry). We would argue that although this can shorten the game length, it is also integral to the game's success -- if the player does manage to rapidly advance the plot through a hunch, that in itself can be a thoroughly rewarding experience. 


\subsection{Outer Wilds}
Outer Wilds is an open world mystery game in which the player explores a solar system trying to understand the natural and artifical phenomena and structures within it \cite{outerwilds}. One of its most central mysteries is also its most important game system: every 22 minutes the sun explodes, destroying everything in the solar system, and the player is transported back in time by 22 minutes. Everything in the cycle repeats again, with only the player acting differently.

The player starts each 22-minute loop on their home planet, where most of the NPCs in the game reside. They have a spaceship which they can use to take off and travel to any celestial body in the system, including six planets, three moons, one asteroid, one sun and several constructed space platforms. Many of these places have elements which run on a timer - for example, one planet is bombarded by rocks from its volcanic moon, and over the course of the 22 minute loop the entire planet's surface is destroyed and collapses into a `white hole' at the planet's core. Many puzzles require being in certain locations at specific times during the loop to take advantage of the state of the simulation.

A major part of solving Outer Wilds' mysteries lies in understanding the history of the Nomai, a race of aliens that lived in the solar system a long time ago. The player is equipped with a tool that can translate their writing, which they find written in walls and on other objects. The player also encounters buildings, ships and devices made by the Nomai and can learn about their function by interacting with them. Nomai writing often reveals information about how different parts of the solar system relate to one another, or how certain unusual features of the solar system work. This takes the form of either explaining things the Nomai built themselves, or explaining discoveries the Nomai made about natural phenomena and how they work.

Whenever new information is discovered through writing, seeing something, or being in a certain place, the game updates the player's log. The log records information and draws connections between related elements, which is retained between loops. For example, discovering a laboratory might create a card representing the laboratory, coloured to show which planet it is on. If the player learns the laboratory was manufacturing something for a project on another planet, a line will be drawn between the laboratory's card and the other project's card. Both connections and cards can be clicked on to show more information, as well as an indicator if there is more to learn in a location. This separates Outer Wilds from the other three games mentioned here, as it does not require the player to demonstrate understanding in order to gain further knowledge (i.e. by entering fates in \textit{Obra Dinn}, translations in \textit{Heaven's Vault}, or specific search terms in \textit{Her Story}).

\begin{figure}
\includegraphics[width=\columnwidth]{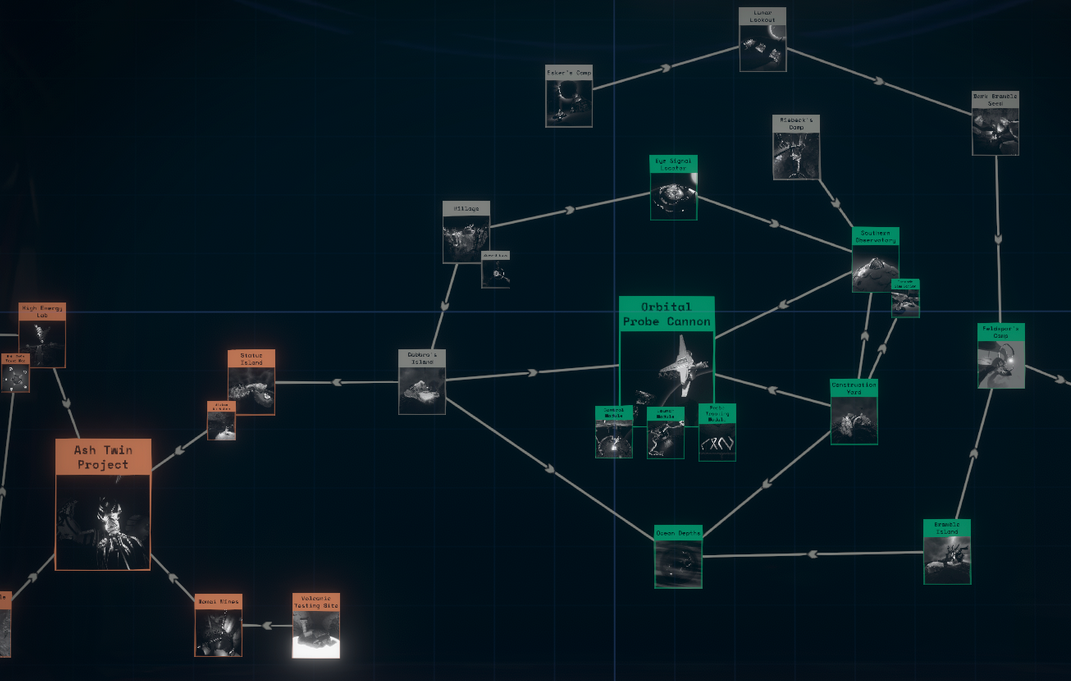}
\caption{The ship's log in \textit{Outer Wilds}. Cards that share the same colour are located on the same planet.}
\label{fig:ow}
\end{figure}

\section{Properties of Information Games}\label{sec:infogames}
In this section we detail some interesting characteristics of information games. This is not a comprehensive list, but the most relevant to the work we present later. This is also not intended to be prescriptive: we do not believe these properties are necessary to qualify as an information game, rather than they may indicate why information games are made the way they are, and what appeal they provide to players.

\subsection{Hard To Model}
Games often need to model the player's understanding of the world or record their progress towards achieving goals. This is usually done in ways which are exactly measurable, often using mathematical or logic-based criteria. In the most explicit cases, a quest demanding the player `Kill 10 Rats' has specifically measurable completion criteria, using discrete values (e.g. quantities and types) and computable conditions (e.g. health points equalling zero). Every step of the process of killing ten rats, from being given the quest, through engaging with the combat system, to the quest being completed, is measurable and modelled within the game. This makes it possible for designers to have the game notice or react to the quest beginning, progressing and ending, or track how well the player is doing in order to help them or hinder them.

Information games are centred around information gathering, processing and application, and often relate this to a mystery of some kind. The information gathered is frequently of a type that is hard for computers to model -- in Her Story, for example, the player will gather information from body language, intonation, idiom and metaphor use, semantics of the interviewee's speech, and more. Return of the Obra Dinn uses knowledge including accents, rank, social class and interpersonal relationships. Heaven's Vault is based on semiotics, communication and language-learning, as well as religion and history. Even Outer Wilds, which ostensibly is about physical systems, is largely about the people who built or studied these systems, how they responded to crises and mysteries, the spirit of discovery and the response to apocalypse.

This drastically changes how information games assess and track player understanding or progress throughout play. Some games completely eschew milestones altogether. Her Story records how many clips the player has viewed, but other than that it does not attempt to model the player's understanding of the mystery -- it simply asks the player when they want to stop. Some games intervene to confirm information, which both helps the player with their investigation and also provides the game with a baseline of what the player must know. The most explicit example of this is Outer Wilds, where the ship's computer log is updated based on various triggers, such as the player entering a particular location. The log will connect locations and highlight important information, regardless of whether the player personally observed or understood this. This can be useful in helping the player avoid getting lost, however overuse of this can also impact the player's sense of understanding and personal discovery.

\subsection{Highly Nonlinear}\label{sec:nonl}
Information game structures are usually very open-ended, allowing players to gather information at their leisure, providing minimal guidance on what the player should do next (see \cref{sec:dist}). Her Story's structure is entirely flat, in that any term can be searched for at any time, and any video clip can be accessed with the right term. Outer Wilds is also flat; the game can be finished in the first cycle of the game if the player knows what to do. Heaven's Vault and Return of the Obra Dinn apply a structure to some of their critical path. Return of the Obra Dinn unlocks new decks of the ship to explore after the player has viewed enough death scenes. Heaven's Vault has a mix: there are many locations the player can visit at any time, exploring side stories, but main story locations are gated behind certain story scenes. Player progress is never gated based on their understanding of the language, however, and the game will still progress even if the player has incorrect translations for important passages of text.

This loose, nonlinear structure serves several purposes. It suits the themes of mystery and discovery that these games have, by letting the player explore on their own, form theories and pursue hunches. However, it also helps mitigate the difficulty of assessing player knowledge, described in the previous subsection. By designing these games to be nonlinear, with open-ended goals, the game has no need to detect the player's progress or understanding. The player can be guided, and can be tested (see \cref{sec:interactions}), but the majority of the game's content is explored at will, which allows the player to set their own pace, and to solve problems in a way that feels natural to them. This helps players pursue aspects of the mystery they feel they best understand, or that feel most promising to them.

\subsection{Distributed Information}\label{sec:dist}
Information in these games is usually distributed thinly across the game, rather than being centralised in a few important scenes. This serves several functions. First, information can be conveyed more naturally, rather than through heavy-handed exposition dumps. There is no specific way one gains a confirmed translation in Heaven's Vault. Instead each new fragment of text, however simple, is an additional step towards confirming new words. Second, it allows for smaller loops of mystery-solving and insight, which helps maintain a sense of momentum. In Outer Wilds the central mystery is solved very slowly, but many short-term mysteries are constantly being solved, contributing to larger cascading solutions to bigger problems or questions. Distributing knowledge also supports the nonlinear structure of these games, which rewards the player for exploring regardless of where they go. There is always something new to be learned, which might combine with some existing knowledge to move the investigation forwards.

In addition to widely distributing information, there is also usually a great deal of redundancy in the information. By this we mean that if an important fact needs to be derived, there are usually many routes through which it can be obtained, and information is often duplicated in multiple locations or contexts to assist in discovery regardless of route taken. For example, in Outer Wilds there are several teleportation devices invented by the previous inhabitants of the solar system. The player can learn about the teleporters in multiple locations by discovering written documents and by interacting with one of several teleporters to test theories about their operation. Some documents are copied and appear in multiple locations (explained in-fiction as scientific reports) while others simply convey the same knowledge but in a different format (such as a written conversation between two people).

\subsection{Simple \& Flat Interactions}\label{sec:interactions}
Player action in information games is usually straightforward, and often does not have a lasting impact on the game state, but instead is more focused on querying the game for information, testing a theory or responding to prompts. Because information games deal with complex ideas that are hard to model (see above) it is also hard for the game to model player interaction with such information. This may be one reason why many information games are about understanding the past, rather than dealing with a current situation, as it removes the player's ability to affect the complex system they are trying to understand. Instead they act more like a researcher, exploring a knowledge base and drawing conclusions from it.

Player actions commonly allow the player to demonstrate knowledge or understanding about some part of the game. In Her Story, the ability to search the video database is completely unconstrained, meaning any English-language statement is potentially valid. However, some queries -- such as the name of the victim, or the manner of their death -- demonstrate that the player has gained enough knowledge to specifically search for a particular term. In some cases the lack of gating can allow players to do this by chance. For example, searching Her Story for a common murder mystery trope can reveal crucial plot information in a way that (arguably) circumvents a more intentional process of understanding. 

Simple interactions also restrict the amount of information the player can extract with a single action. This means that the player generally only derives basic knowledge, and the task of assembling that knowledge correctly must be done by them. This is vital to maintaining the sense of mystery and discovery, because it keeps the player's more complex ideas about the mystery unconfirmed for longe. For example, in Heaven's Vault a player might correctly translate most of a sentence but mistranslate a key noun in such a way that it changes the interpretation of the historical event being described. Being able to maintain these misunderstandings means the player gets to correct themselves by finding contradictory information later, rather than being simply told they are wrong during a validation check.

\subsection{Environmental Storytelling}
\textit{Environmental storytelling} is a broad term used to describe the conveyance of narrative information by constructing a scene for the player to uncover or explore. Game designer Bart Stewart describes environmental storytelling as `the art of arranging a careful selection of the objects available in a game world so that they suggest a story to the player who sees them' \cite{stewart}. Environmental storytelling is often supplementary, rather than being used to convey plot-critical information. Although still in common use, its overuse in some contexts has led to it being stereotyped. In particular, the use of skeletons in various contexts has become a running joke. As game designer Ben Esposito put it: `in game design, ``environmental storytelling'' is the art of placing skulls near a toilet' \cite{esposito}.

Some information games do not have the player inhabit a physical space (such as Her Story) but games which do tend to leverage environmental and other indirect storytelling forms in order to scatter extra clues and information for the player. \textit{Outer Wilds} uses a technique similar to audio logs by having the player find writing from the Nomai strewn across the solar system. Not all of the text is critical to solving the game, and some of the information is duplicated to make it more likely that the player will find it. \textit{Return of the Obra Dinn}'s death scenes are a form of environmental storytelling, where the arrangement of people and objects imply a story, often with hidden details that need to be uncovered by careful inspection.

The use of environmental and indirect storytelling techniques is prominent in information games because it provides a way to reward observation, perception and exploration, all of which are closely related to the games' themes of knowledge acquisition and use. Information can be conveyed simply by the relative positions of objects, the poses of characters, or the presence of something that is out of place. They are also good methods for conveying information in small amounts (see \cref{sec:dist}), and in non-linear games where the player may take many paths through the game.

\section{Motivation}\label{sec:motivation}
Procedural content generation (PCG) and related terms like `generative software' are broad terms used to describe a class of algorithms that encode a process for producing something. Often these algorithms are designed to have a large space of outputs, but this is not always the case. Some procedural systems exist to automate repetitive processes at scale \cite{autocode}, or to produce a single large and complex artefact from a comparatively small amount of code \cite{genart}.

When applied to games, procedural generation is often used as a source of variation or unpredictability. This is used in many different ways, including achieving goals in visual art  (producing many small variations of plants and trees to create convincingly natural spaces \cite{speedtree}) as well as gameplay effects (such as Yu's desire to avoid rote learning of platformer levels \cite{spelunkybook}). Marketing campaigns for games that use procedural generation often leverage the scale of their generators as an indicator of value or quality (e.g. Borderlands 3 declaring their generator could create over 1 billion guns).

Our initial motivation for this work was to design games which eschewed classical PCG aesthetics \cite{smith}. We were interested in particular in two common trends in PCG: the attitude of treating generated content as throwaway, and the idea that players learning how generators work is undesirable (as it makes them predictable, undermining their role as a source of surprise). In the first case, we were interested in what we call \textit{small-batch PCG} in which the player becomes invested in a single generated output and is encouraged to explore and understand it. In the latter case, we were interested in the idea of designing a game in which the player is encouraged to learn the PCG system's patterns and behaviour.

As we began planning our prototypes, the connection to information games became clearer. The idea of encouraging the player to engage deeply with a system fits well with the format of an information game, in which the player's progress is linked to their understanding of an artefact of some kind. This both encourages engagement with a single piece of content (for one playthrough) but also encourages the act of understanding the generative system, by making this part of the process of understanding and solving a mystery.

PCG also allows us to examine the impact of variation and surprise on information games. In much the same way that this was used to change player experience in traditionally static genres like platformers \cite{spelunky}, shooters \cite{nuclearthrone} and survival games \cite{minecraft}, we wanted to investigate how PCG could change information game design, both in terms of repeated playthroughs by the same player, and variation of playthroughs across a community of players. 

\section{Nothing Beside Remains}\label{sec:nbr}
\textit{Nothing Beside Remains} was developed for the 2018 Procedural Generation Jam. The game is played as a top-down, grid-based game in the style of a roguelike, with Unicode characters representing everything in the world, in the style of ASCII roguelikes. We used a two-phase generation system to first simulate a small village using an abstract model, and then render it in the game world for the player to explore. The village is simulated until it collapses, and the rendering process builds a ruined version of the village that is adjusted to convey a sense of how the village came to meet its demise. In this section we will describe how the world is generated, and how gameplay is structured. It can be played online in a browser\footnote{cutgarnetgames.itch.io/nothing-beside-remains}.

The following text is shown to the player before play begins:

\begin{quote}
\textit{You've arrived in a ruined village lost in the desert for hundreds of years. Explore the village, see what remains of it, and see what you can learn about the people who lived here. What did they believe? What caused their downfall? You may never know for sure, but perhaps there's a clue or two left in the sand...}
\end{quote}

NBR's game structure is open-ended, with no evaluation or fixed goal for the player. Instead, we invite the player to explore as much of the world as they want, gather information, and quit whenever they feel like they have seen enough, in a similar approach to Her Story. While we could have simply asked the player what they thought was the cause of the village's collapse, we felt that this would undermine the broader understanding the player was developing of the village's history in general, not just how it ended. Judging by feedback from players, discussed later, we believe this decision was justified.

\begin{figure}
\includegraphics[width=\columnwidth]{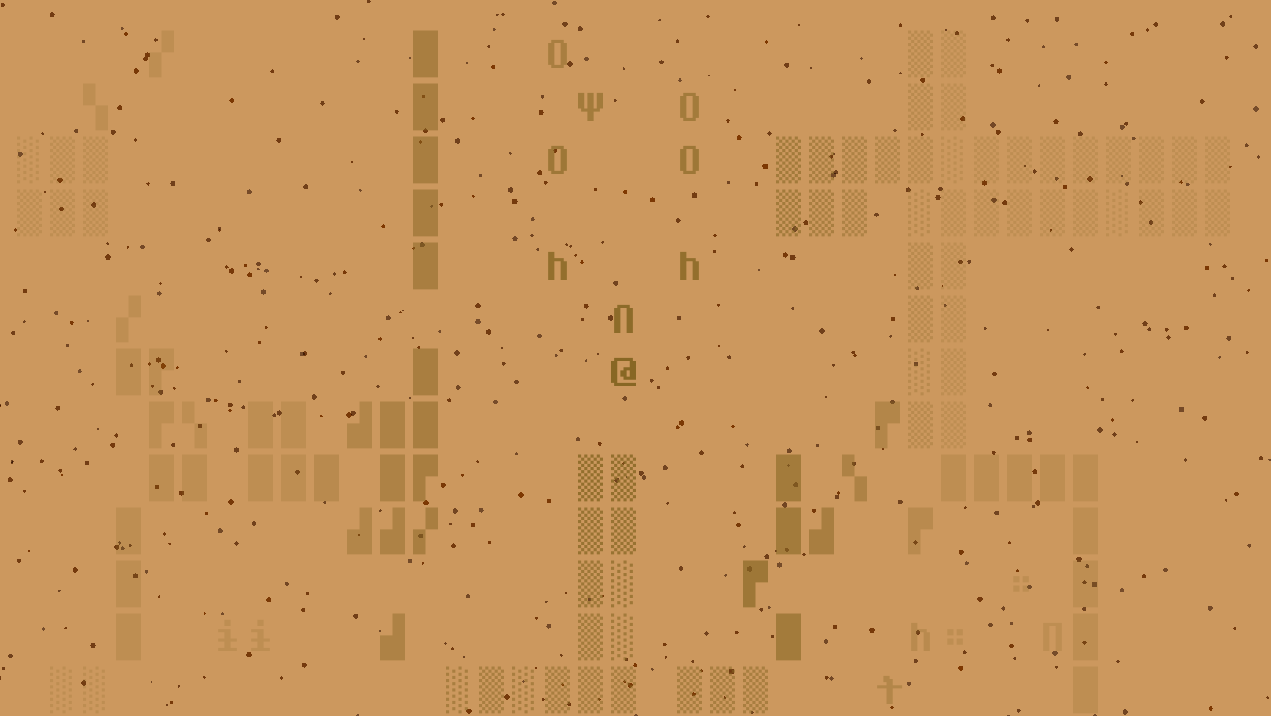}
\caption{A screenshot from \textit{Nothing Beside Remains}.}
\label{fig:nbr}
\end{figure}


\subsection{World Simulation}
There are two phases to village generation: an abstract simulation phase, and a rendering phase. This structure is inspired by games like Dwarf Fortress which simulate their world in the abstract and only render things explicitly that the player encounters (so events from history are not modelled in detail, for instance). In the simulation phase we build a high-level model of the village and simulate it until we meet one of several ending conditions. There are two key components to the model: the \textit{ecosystem}, which models the health of various aspects of the natural world, and \textit{society}, which models the village itself. 

The society model is the simpler of the two: it records the number of people living in the village, how many are of working age, how much food is in storage, and then some cultural features such as what materials they crafted with, what numbers they held to be special, what flowers they cultivated, and (after the simulation) records the reason why the village collapsed. The ecosystem model is slightly more complicated. It tracks three key features: the average temperature, the density of hostile fauna in the area, and the health of the ecosystem (i.e. how well it can support agriculture). Each of these features has a current value, a minimum and maximum cap, and a cap on how much the current value can change per simulation tick (to avoid extreme shifts).

Once these two models are initialised, we run a turn-based simulation of the village, with each turn stepping through the same sequence of checks. Temperature and hostile fauna fluctuate randomly. The ecosystem then has a chance to take damage, with the chance increasing with temperature. If the ecosystem health reaches zero, the simulation ends. The population goes down based on the intensity of hostile fauna. If the population reaches zero, the simulation ends. After this check, population increases slightly based on the crop surplus. Crops then grow or shrink based on temperature and population size. If the crops stock reaches zero, the simulation ends.

This simulation has three possible endings: ecosystem collapse, overrun by predators, and famine. The parameters of the simulation, such as the starting values for each variable and their caps/drift rates, were set experimentally in order to achieve an approximately equal distribution of the three endings, which we verified by running 500 simulations and recording the balance of outcomes, then tweaking parameters to affect ending probabilities. The remaining parts of the model, the cultural features of the society like favoured crafting materials, are set randomly as they only impact the final rendering the village.

\subsection{World Rendering}
Once the simulation has ended, we construct the game space for the player to explore. We use a 100x100 grid of tiles, which we split into 10x10 regions which we use to assign larger features to. Then we add several features which are always present in every generated village. The first are water sources -- we add a number of lakes to the map, the quantity and size of which are based on on final ecosystem health. Large lakes or swamps suggest high ecosystem health at the end of the simulation, while smaller and fewer lakes might imply the village collapsed because the ecosystem collapsed (or was simply close to collapse when another fate befell it).

Next, we add two features which are common to every village: the place of worship and the statue. The statue is inspired by the Shelley poem \textit{Ozymandias}; it has collapsed into fragments, and a plaque at its base, which the player starts in front of, has an inscription describing the ruler of the region. Figure \ref{fig:nbr} shows the collapsed statue. We procedurally generate the inscription text (see below) and the scattering of statue pieces.

The place of worship is situated near the statue, and is a long building taking up a 20x10 space. It is laid out somewhat like a Christian chapel might be -- there's a large area with rows of pews, and then an open space at the end with an altar.

After the fixed points have been added, we lay out a path from the worship hall to the nearest water source, and randomly add houses to the edge of the path. This forms the `inner' village. Once this is complete, we add secondary roads branching off this main road, and keep adding houses to the roads -- any spot adjacent to a road that does not contain a building has a chance to branch off a new road. Each house added onto these secondary roads has a chance to be a barn or a field instead, with the chance increasing as the road gets further from the center of town. This means that farmland tends to be kept on the outskirts, with more houses closer to the place of worship. We continue this process until a random number of roads have been added, or there are no more blocks to add roads onto.

\subsection{Decorating Rooms}
The outer structure of every building (including fields and places of worship) are laid out next. We first lay the structure out as it would have been originally built, and then randomly destroy or displace parts of the structure to simulate damage and decay. The chance of damage is related to temperature levels, to simulate harsh weather. We also adjust visual representation (e.g. walls become broken into small rocks). This not only conveys one of the simulation factors visually, but also helps convey a sense of ruin, as the player can move through broken walls and fences to traverse the village.

After laying down the outer structure, each building is populated with items based on its type and environmental factors. Each building has a set of items it pulls from with a specific probability. Houses contain items such as tables, chairs, cutlery, children's toys; fields contain various crops and wild plants; both barns and fields might contain animal skeletons. The chance of some items appearing are affected by the simulation - children's toys are influenced by the birth rate and general population growth; if there are very few children's toys this might suggest the village ended with the population being wiped out. Crops, and plants in general, will be less common in fields if crops failed or the ecosystem collapsed, while weeds will be more common. Throughout all buildings, predator skeletons will be more common based on the hostile fauna density. They have slightly different appearances to cattle skeletons in-game, and different descriptions too.

\subsection{Generating Descriptions}
Any object in the game can be examined to read a description. Some are static, while others have generative elements. There are two layers to our generative text system. The first is a Tracery layer which uses standard syntax for generating text \cite{tracery}. Within these grammars are markers for a second layer of dynamic text, which is handled by our system after Tracery has finished generating. These markers refer to elements which are dynamic, but consistent across all text within a particular instance of the game. 

For example, generated descriptions which want to mention the material something is made out of will leave a dynamic marker `@MATERIAL@' in the Tracery output, which the system will later replace with the material the village used most for crafting. This two-phase text generation allows us to mix dynamic elements into the procedural descriptions. Most of these dynamic elements are decorative, but one element in particular is very important: `@DREAMENGRAVING@', which describes the village's hopes and fears, and is specifically linked to how the village ended in ruin. This is found in the descriptions of engravings in the worship hall. One possible engraving is of a lush forest scene. This means the village in this game suffered from an ecosystem collapse. This is the only piece of information in the game that unambiguously confirms the fate of the village, although the player does not know this.

\subsection{Gameplay}
The player moves around the world using simple directional movement, and can inspect almost any object they find by bumping into it. There are no combat encounters, puzzles, or challenges (other than trying to discover knowledge). Figure \ref{fig:nbr} shows a screenshot from the game. The game has no specific ending; instead the player is encouraged to explore as much as they want, and quit when they feel satisfied. We also recommend to the player that they only play the game once.

\subsection{Evaluation}
\subsubsection{Player Feedback}
By far the most satisfying outcome from this prototype was the feedback from players. In particular, several players wrote responses to the game, some styled as a kind of archaeological report on what they had found. These included descriptions of the village they had explored, as well as theories about the people who had lived there. For example, one player wrote: ``The evidence isn't conclusive, but it appears that this village lost its ability to grow crops through a change in the environment, with water drying up and plants dying off.'' 

Of course, it is hard for us to know whether this player's assessment was accurate, or if they simply misinterpreted the amount of water remaining. This is one of the motivations for advising players to only play the game once, because it stops them from being able to resample the generator and learn what the limits are on the generator's variation. Instead of seeing many villages and learning what the biggest and smallest lakes are, the player is instead forced to relate features such as the size of a lake to real-world analogues, as they have nothing else to compare it to. We believed this would encourage players to engage more deeply with the world, despite it being procedurally generated.

A second side effect of asking the players to only play once is that they are unable to gauge which parts of the world are simulated and which are static. For example, the same player remarked that ``religion was likely a daily part of their lives, as several houses had altars with offerings of perfume''. However, repeated playthroughs of the game would reveal that perfume always occurs inside houses with the same probability. Interestingly, some features that were dynamic between playthroughs can appear less meaningful when the game is only experienced once. Each society has a sacred number, and one way this manifests is in the design of chairs, which always have a sacred number of legs. One player reported: ``One quirk that I found particularly amusing was the number of three legged seats, I think just about every stool or chair I encountered was missing exactly one leg''. Three legs did not seem significant, and in fact seemed strange or arbitrary. Another player encountered a different sacred number, which was more of an outlier, and interpreted it more accurately: ``Chairs typically had seven legs, which almost certainly means that the number seven had cultural, probably religious, significance for the villagers.'' Of course, the nature of history and archaeology means theories cannot always be proven `right' and thus the idea that these chairs were simply missing a leg is a valid hypothesis. What is interesting here is that because only one village was explored by each player, a dynamic feature was interpreted very differently by each player, leading to completely different interpretations.

\subsubsection{Abstract Simulation}
Abstract or high-level simulations scale better than detailed simulations, and can be simulated faster. Even with abstracted simulation, generating a world in Dwarf Fortress can take many minutes. Even though our prototype is small by comparison, the approach will scale well to larger simulations. Abstraction makes sense in many generative contexts because the player is mostly concerned with the end result of the simulation. Although the history of a generated world in Dwarf Fortress can have relevance and interest to the player, most of the fine details do not impact the player's objectives (for example, the path through a village a farmer took eight hundred years ago to fetch water). 

Although abstract simulations have their benefits, we feel that from the perspective of information games they had a significant drawback, namely that the rendering has too indirect a relationship to the simulation. A simulation in NBR effectively defines a \textit{space} of renderings, which is sampled by the generator in the rendering step. The problem is that the process of generating a rendering makes decisions about the world that do not have a basis or justification in the simulation. For example, the number of farms, their position relative to the water sources, and the quantity of houses have no relationship to the population of the village or its food stocks. However, as discussed in previous sections, information games encourage the player to interpret the environment they explore, which means players end up drawing conclusions that the simulaton cannot back up.

For an aesthetic or tonal piece, this is not a particular disadvantage, and in fact seems to have helped some players get a lot more out of the game. One player commented on the presence of a trident in the statue, wondering if it related to a past as a fishing village, while another hypothesised that the area was underwater at one point. These colourful interpretations of extremely small (and random) details are encouraging, and as an open-ended exploration game with a creative writing twist we find it to be quite effective. However, for information games that attempt to validate the player's knowledge (for instance, if we had asked the player to infer the fate of the village) it might be frustrating to have so much information be disconnected from the game's model of what happened.

We believe that abstract simulations do work for generative forensics games, as long as the player is not rigorously evaluated on their understanding of the system. A game with the structure of \textit{Elegy for a Dead World}, for example, where players are encouraged to complete creative writing tasks in response to exploring worlds, would be an excellent way to incorporate a personal response to a space generated in this way. However if the player was expected to build on and be motivated by their knowledge (as Francis suggests in his original proposal of the term information game) then this approach leaves too many loose ends that the game cannot tie up.

\section{Condition Unknown}\label{sec:cu}
\textit{Condition Unknown} (CU) is a second generative forensics prototype, developed one year after \textit{Nothing Beside Remains}. After reflecting on the response to Nothing Beside Remains, we decided to explore a different approach to generative forensics, this time working with a much more detailed simulation, and setting the player a simple goal. In CU we use an agent-based simulation to model a science-fiction story about a catastrophe at a research station. The player arrives after the event, and is tasked with identifying the station crew and recording their fates, inspired by Return of the Obra Dinn.

The following text is shown to the player before play begins:

\begin{quote}
\textit{You've arrived on the outskirts of a remote research station that was excavating an unusual find, deep into the Arctic wastes of this planet. Explore the station, discover what happened, and document the fate of the crew that lived there. }
\end{quote}

\begin{figure}
\includegraphics[width=\columnwidth]{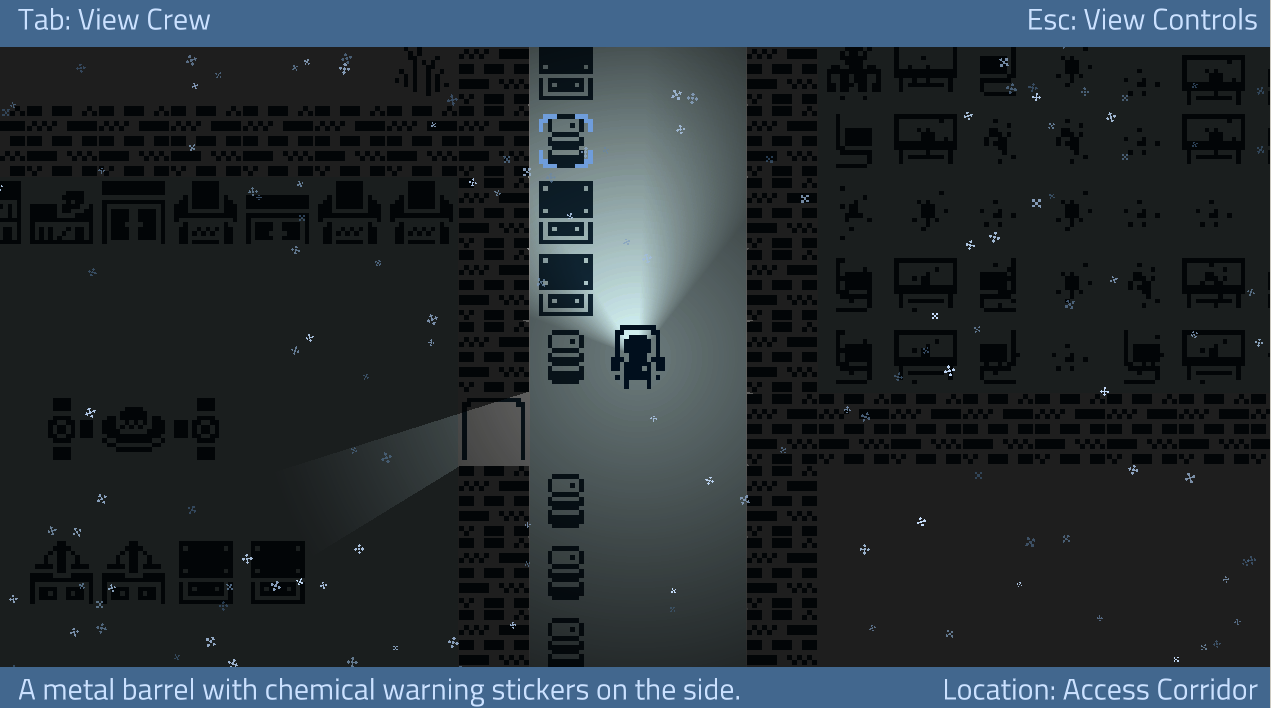}
\caption{A screenshot from \textit{Condition Unknown}. The player is in the center of the screen, shining a torch in the direction of the mouse cursor.}
\label{fig:cu1}
\end{figure}

The player does not have to perform this task, and can simply explore the station to piece together the story, much like NBR. However, if they wish to they can identify causes of death and submit a final report, which ends the game. The player is then told how many of their responses were correct. In this way, CU experiments with a more objective-focused approach to information games, closer to Return of the Obra Dinn than Her Story. As with NBR, there are no combat encounters or challenges, other than the optional challenge to identify crew deaths. Figure \ref{fig:cu1} shows a screenshot from the game, and it can be played online in a browser\footnote{cutgarnetgames.itch.io/condition-unknown}.

\subsection{World Simulation}
Every simulation in CU is different, but has the same core structure and premise. A research station is attacked by an `anomaly' during experimentation. The crew try to fight it or escape, but all of them fail and are eventually killed somehow. World generation has two phases: a construction phase where we set up the initial conditions for the simulation, and a simulation phase which begins at the point the anomaly attacks and stops when everyone in the station is dead.

\subsubsection{Construction Phase}
In this phase we build the basic structure of the research station, place all characters in their starting locations, and check initial conditions. First we place a random number of corridors down, all connected to each other, and then we place a random number of rooms attached to the corridors, making space for doorways. We then assign room types: the most important room is the station entrance, which is always the southernmost room, and is the only room that starts with an exit door into the outside world. Other room types are assigned randomly. There is always exactly one of the following rooms: Mess Hall, Residences, Laboratory and Security Office. All remaining rooms become secondary laboratories.

We then assign scenery to each room. Whereas NBR had a flat chance to place scenery in any space, in CU we use grammars to ensure we place scenery in patterns that make more sense. For example, a desk will have a chair in front of it. A custom scenery placement system loops through the room, placing certain kinds of scenery down depending on room type, the available space and whether or not it will be adjacent to a wall or doorway. We ensure scenery is never placed in front of a doorway, and every pattern leaves a 1-tile gap around it to guarantee that the entire station remains walkable. This was less important in NBR because the player did not need to explore everywhere, but in CU navigation is vital both for the player to discover clues, and for crew members to have free movement during the simulation. 

Between five and six characters are then generated. Their two main properties are their name -- drawn from a pool of 15 hand-authored names -- and their profession. There is always one Security Officer, one Logistics Officer, and the remaining staff are all Scientists. We place characters in the station based on their job: scientists start in one of the labs, security guards always start in the security station, and logistics officers always start in the mess hall. Their profession affects their description, which helps the player identify bodies later. Finally, we add the last character, the creature attacking the lab. This is referred to in the project as `the anomaly'. The anomaly always begins the simulation in Lab \#1. 


\subsubsection{Simulation Phase}
Once the world has been constructed, we begin simulating the events that led to the destruction of the station. This is done using a turn-based system in which the station staff, the anomaly and any dynamic objects all take turns. Dynamic objects are non-intelligent entities such as fire, which has a percentage chance to spread to nearby tiles, as well as a chance to extinguish by burning out. Dynamic objects also respond to events, such as fuel barrels exploding if shot, or walls being destroyed by explosions.

The anomaly's AI system is the next simplest. The anomaly moves towards any human it is targeting, as long as it can see it still. If it has no target and sees a new person, it will chase them instead. Finally, if someone attacks it, it will turn and target them. If the anomaly begins its turn adjacent to a human, it will kill them (they combust, the cause of death is burning). If too much time has passed since it saw a target, we simply provide the anomaly with the location of a random crew member (we discuss this later). It also has a small chance to set the tile it is on on fire, and will set fire to a larger number of nearby tiles if it is shot.

The station crew have the most complex AI. The AI uses a priority-based series of checks to respond to events based on their severity, as well as the current state the crew member is in (what they are doing, what other things are happening to them, and so on). For example, seeing a dead body is a high priority event that will trigger a response and make them stop what they are doing. However, if they are already doing something urgent (such as fleeing the station) they may not regard the event as being as important. Events can be triggered by what the crew member sees (e.g. fires, dead bodies, anomalies), hears (e.g. explosions, screams, gunshots) or knows (e.g. has been told about the anomaly, has been told someone has died). 

We do not have space here to exhaustively describe the web of priorities and events, however we can provide an overview by talking about the narrative structure the crew's AI was designed to tend towards. Although the initial conditions and the simulation itself are fairly unconstrained, we designed the AI systems with two important `points of no return' that function as barriers between different acts in the narrative. Broadly speaking, these acts are:

\begin{itemize}
\item 1. Opening: no-one is alerted, crew working as normal.
\item 2. Panic: station is alerted, crew seeking shelter or fleeing.
\item 3. Climax: crew formulate a plan to resolve the situation.
\end{itemize}

The transition from the first to the second act is not uniform -- crew members transition at different times depending on how knowledge of the anomaly spreads through the station. Each crew member has a panic level that is increased by witnessing or being told about certain events. Seeing a fire is a cause for concern and will cause the crew member to seek shelter in another room, for example, but it does not tell them specifically about the anomaly. If their panic reaches a certain level, they will attempt to find a safe place to meet up with other crew members. There are two event sequences that commonly occur to bring the entire station to full alert: the first is a crew member witnessing the anomaly and surviving long enough to radio their sighting in to the station. The second is the crew hearing something suspicious, at which point the security officer will go to investigate, which leads to them witnessing the anomaly and alerting the station.

Once the station is alerted, crew members run to rooms far away from reported danger. There are many events that can trigger here, because movement often triggers new sightings of other dangers, including dead bodies, and they can also meet living crew members too. In addition to this, because the crew members have imperfect information they may attempt to flee to locations which are also dangerous -- for example, fleeing a fire and running into a room with the anomaly. We prevent the crew from leaving the station by having it set in a cold and inhospitable environment, where leaving would kill them. This means that over time safe areas become smaller, and the anomaly will eventually track down crew members if it does not encounter them naturally. The transition to the third act is fixed, and always triggers when there are only two crew members left.

During the third act, the remaining crew members pick from a set of `endgame' plans for how they intend to kill the anomaly or escape the station. There are two possible endgames currently in CU: in one, they raid the security office for weapons and confront the anomaly; in the second, they run outside and attempt to flee the station. Both plans are guaranteed to result in their death. If any crew member spends more than five turns outside the station they die of exposure, and the anomaly cannot be killed so confronting it will result in it attacking them. When all crew members are dead, the simulation terminates, and the anomaly is removed from the game.

\subsubsection{Message Passing}
Although the final state of the station does impart some information about what happened to the crew, it is not enough to identify the crew, and chronology in particular becomes difficult to put together because events may overlap (for example, fire may spread through an area twice). To provide additional information to the player, as well as to convey a sense of a real sequence of events taking place and tell an interesting story, the crew's AI doubles as a messaging system in which they radio each other with reports about events and actions. After the simulation is complete, every recorded message is assigned to a \textit{terminal} -- a special world object that appears all over the station. We place one message in each station, with the most chronologically early messages placed nearest the entrance, and then later messages being placed in terminals deeper into the station. This is a simple attempt to stagger `exposition' so that later messages are found as the player sees more and more of the station's destruction.

Messages fulfil several roles in the design. Most obviously, they contribute to the tone or atmosphere in some way, adding small details to the narrative. We partition the information content of the messages into three types: reports, intentions and updates. Reports describe an event that a character has \textit{witnessed}, such as finding a dead body. Intentions describe what a character \textit{plans} to do next, such as going to a particular room to find shelter. Updates are a special kind of message where we force two characters to exchange information somewhat artificially in order to help fill in gaps in the player's knowledge later. For example, when a lot of action is occurring at one end of the station, characters in the safe end may have no reason to send messages about their whereabouts. To balance this, characters sometimes mention their location when in dialogue with other characters, usually in the form of a simple `Where are you now?' query or a `I'm in <location>, get to us if you can!' request as part of another update.

Every message, regardless of its content, is also marked with a timestamp and the name of the person sending the message (some messages also include responses from other people, but we only explicitly provide the first person's name). The timestamp is calculated as an offset in minutes from a random start time, based on the number of turns that have elapsed. As an example, if the random start time is 10:41 am, a message sent on turn 10 will have the timestamp 10:51 am. This allows the player to put messages in context, and also allows important deductions to be made about the state of certain crew members. In this way, even a message that contains no useful content in its message still conveys important information, in that it confirms the sender of the message was alive at the given timestamp.

\subsection{Gameplay}
Similarly to NBR, the player moves around the grid using simple directional movement. We made some small quality-of-life improvements by showing descriptions of items on mouseover, rather than forcing the player to walk up to the object, for instance (with the exception of terminals which must be read by touching them). We also improved the rendering, with sprites for objects and a lighting system. The player has a torch they can shine to reveal parts of the station, which helps them discover things more gradually, as they can only see things in the room their character is currently in.

\subsection{Evaluation}
\subsubsection{Player Feedback}
NBR's presentation to the player emphasised exploration and discovery, even framing the player as a historian or archaeologist of sorts. There was no explicit objective, and we instructed the player to explore at their own pace and quit when satisfied. With CU we gave similar instructions, but also provided an optional task. In addition to this, the game is built around a more specific event in a shorter time period, with less ambiguity and more first-hand information to read. As a result, we found that response from players was less creatively interpretive than before, and more focused on the task. Many players reported their success rate on the final task to us, but none retold the story of their playthrough, or attempted to retell the events that took place in their station. In both cases our playerbase is small -- at the time of writing, CU has had 500 players in three months, while NBR has had 1,000 players in a year -- but we suspect the more task-focused nature of CU is an important factor in the changed response.

General feedback to CU was positive. Although we recommended to players that they did not replay the game, many reported that they enjoyed doing so, and felt that it was more enjoyable to replay compared to NBR because the purpose of the game \textit{was} to understand the generator, and effectively become more skilled at investigating and solving mysteries. However, another player noted: ``on the second playthrough... having figured out the clockwork of [the generator], the task seem[ed] much more mathematical''. This matches our expectation that players rapidly adapt and understand what the generator is responsible for, but it seems players differ in how much this bothers them.

\subsubsection{Robustness of Deep Simulation}
Although one might assume that finer-grained simulation would lead to less flexibility, we found that due to the specific design of CU our simulation was very robust to certain kinds of failure. When developing agent AI for characters the player can see, it's important to ensure the AI does not do anything that would break the illusion of being a real thinking, feeling human being. YouTube is littered with clip reels of AI ignoring obvious signs of danger, walking into walls or off cliffs, or falling for simple traps. However, in generative forensics games the player only sees the end result of the simulation, and all other information they glean is information we have chosen to release. This allows us to develop AI agents which are more loosely specified and that make mistakes, as long as those mistakes don't affect the integrity of the information left behind.

For example, the exact route a character takes from one room to another in CU is not seen by the player. We use basic A* to path between locations, which means characters often take more dangerous routes because they are more optimal in terms of distance travelled. Although we forbid A* to path through tiles which are on fire, we don't discount tiles \textit{adjacent} to fire, which are dangerous as they may catch fire when the character is on them. This leads to characters dying unnecessarily in fires. If the player were able to see the precise route they took, this would make the characters seem less intelligent, less human and less engaging. However, in CU the player only learns about this event through what we choose to report: the character might report they are leaving to move through the facility; another character may report seeing their body; and the player can discover the character's final resting place. At no point is the player exposed to the problems with their pathing algorithm.

This can be seen as an invitation to write worse AI, but in fact it should really be seen as an opportunity to write \textit{better} AI. This design structure allows us to vary our focus and level of detail in our agents, meaning we can put extra effort into writing AI to handle social situations or knowledge representation, and worry less about the exact strategy an agent uses when fighting a monster or pathing through a burning building (if we know the player will only learn about the event's outcome). This gives us a chance to focus on different things, and gloss over the rest with good writing and presentation. This is similar to the idea of \textit{story sifting} introduced by Ryan in \cite{ryanthesis}, but rather than identifying potential story templates in a trace of events, instead we simply choose not to report events which do not contribute to the story (however, in the case of CU this is baked into the system, rather than being a decision an AI narrative manager must make on-the-fly).

\section{Future Work}\label{sec:fw}
Both of our prototypes led to useful feedback and raised more questions about what generative forensics can do for game design and how these games can be best designed. Many points of future work exist, but we outline a few prominent lines of inquiry here.

\subsubsection{Knowledge Guarantees}
We explicitly avoid trying to reason about whether Condition Unknown's core mystery is solvable at generation time, although it would be fairly straightforward to do so. There are many procedural generators which try to make firm guarantees about their output, many of which involve ensuring there is a valid solution to a problem. We decided not to do this for our prototypes: we wanted each output to not necessarily be solvable, to encourage players to consider when they have exhausted all avenues of investigation, and to reflect the fact that sometimes investigations or archaeological surveys do not yield answers.

However, this is a departure from traditional information games. Many allow the player to end the game without understanding everything, but all of them guarantee that the truth can be known if the player explores and understands enough (with the possible exception of Her Story which maintains an element of ambiguity for some plot points). We believe that games which do not guarantee a solution are interesting and valuable, but it is also worth exploring how we can build generative forensics games that guarantee solvability too. This might initially involve quite clumsy solutions, for example by hiding unambiguous details in every station in Condition Unknown. It might also be possible to write solvers or reasoning systems to play the game, but this might encourage the games to move further towards more easily-modelled knowledge, which arguably departs from the spirit of information games.

\subsubsection{Alternative Evaluations}
Information games usually either do not evaluate player knowledge, or do so in a fairly strict, unambiguous manner (see \cref{sec:infogames}). However, we were struck by player responses to Nothing Beside Remains, and how performing a creative act in response to gameplay (in this case, writing a report on their findings) seemed to spur a deeper engagement with the game, as well as encouraging a different response than games normally ask of the player. A similar experience can be found in \textit{Elegy for a Dead World} in which the player completes creative writing tasks in response to exploring hand-designed planets.

Incorporating creative writing tasks as part of the explicit response to the game is an interesting design extension that embraces the unpredictable and varied nature of generative forensics games. Although the game itself cannot understand or respond to such work, we envisage a meta-level community of players who write reports on their experiences, and collectively gain a better understanding of how the generator works simply by reading the reports written by other players. This would be an interesting way to encourage community discussion about the game, and would also add the potential to explore social game design ideas as espoused by designers like Dan Cook and Tanya Short \cite{cook, short}.

\section{Conclusions}\label{sec:conclusion}
In this paper we examined the category of games known as `information games', in which players solve a mystery through exploration, understanding and reasoning, and suggested a new class of game, `generative forensics games', in which the player is trying to understand the output of a generative system. We described two prototypes that explore this idea from different perspectives, \textit{Nothing Beside Remains} and \textit{Condition Unknown}. We motivated and described their design, discussed their evaluation, and pointed to future considerations for work on such games in the future.

Procedural generation is often seen as a way to obtain `more unpredictable stuff'\cite{mus}, but as a design tool it has many different uses, and combining it with unfamiliar design structures can be enlightening and entertaining. We believe generative forensics games are a promising design space to explore, and may also point to a need for better tools to help designers build, understand and work with generative systems \cite{danesh}. Being able to use generative systems as a natural part of the design process, as one might play with a physics system or edit a dialogue tree, is necessary to open up this design space to more people. We look forward to working more in this area in the future.

\section{Acknowledgements}
Thanks to Andrew Dunn for naming Condition Unknown, and to Tom Francis for discussions about information games.

\bibliographystyle{ACM-Reference-Format}
\bibliography{biblio}

\end{document}